\pdfoutput=1

\documentclass[11pt]{article}

\usepackage[final]{acl}

\usepackage{times}
\usepackage{latexsym}

\usepackage[T1]{fontenc}

\usepackage[utf8]{inputenc}

\usepackage{microtype}

\usepackage{inconsolata}

\usepackage{graphicx}

\usepackage{amsmath}
\usepackage{amssymb}
\usepackage{mathtools}
\usepackage{amsthm}
\usepackage{bbding}

\usepackage{microtype}
\usepackage{subfigure}
\usepackage{booktabs} 
\usepackage{multirow}
\usepackage{array} 

\usepackage{colortbl}
\usepackage{makecell}
\usepackage{amsmath}
\usepackage{algorithmic}
\usepackage{algorithm}

\providecommand{\cref}[1]{Chapter~\ref{chap:#1}}

\providecommand{\mat}[1]{\ensuremath{\boldsymbol{#1}}}

 \providecommand{\mH}{\mat{H}}

\providecommand{\mW}{\mat{W}} \providecommand{\mX}{\mat{X}}

\title{Achieving Binary Weight and Activation for LLMs Using Post-Training Quantization}

\author{
 \textbf{Siqing Song\textsuperscript{1,2}},
 $\textbf{Chuang Wang\textsuperscript{1,2}}^\ast$,
 \textbf{Ruiqi Wang\textsuperscript{3}},
 \textbf{Yi Yang\textsuperscript{3}},
 \textbf{Xu-Yao Zhang\textsuperscript{1,2}}
\\
\\
 \textsuperscript{1}MAIS, Institute of Automation, Chinese Academy of Sciences, China
\\
 \textsuperscript{2}School of Artificial Intelligence, University of Chinese Academy of Sciences, China
\\
 \textsuperscript{3}Central Media Technology Institute, Huawei Technologies Ltd.
\\
    \small{
        songsiqing2023@ia.ac.cn,  \{chuang.wang, xyz\}@nlpr.ia.ac.cn, \{wangruiqi12, yangyi16\}@huawei.com
    }
}

\begin{document}
\maketitle
\begin{abstract}
Quantizing large language models (LLMs) to 1-bit precision significantly reduces computational costs, but existing quantization techniques suffer from noticeable performance degradation when using weight and activation precisions below 4 bits (W4A4). In this paper, we propose a post-training quantization framework with W(1+1)A(1×4) configuration, where weights are quantized to 1 bit with an additional 1 bit for fine-grain grouping and activations are quantized to 1 bit with a 4-fold increase in the number of channels.   For weight quantization, we propose utilizing Hessian-aware fine-grained grouping along with an EM-based quantization scheme. For activation quantization, we decompose INT4-quantized activations into a 4 × INT1 format equivalently and simultaneously smooth the scaling factors based on quantization errors, which further reduces the quantization errors in activations. Our method surpasses state-of-the-art (SOTA) LLM quantization baselines on W2A4 across multiple tasks, pushing the boundaries of existing LLM quantization methods toward fully binarized models. 

\end{abstract}

\footnotetext{$^\ast$Corresponding author.}

\section{Introduction}

\begin{figure}[t]
\vskip 0.2in
\begin{center}
\includegraphics[width=\columnwidth]{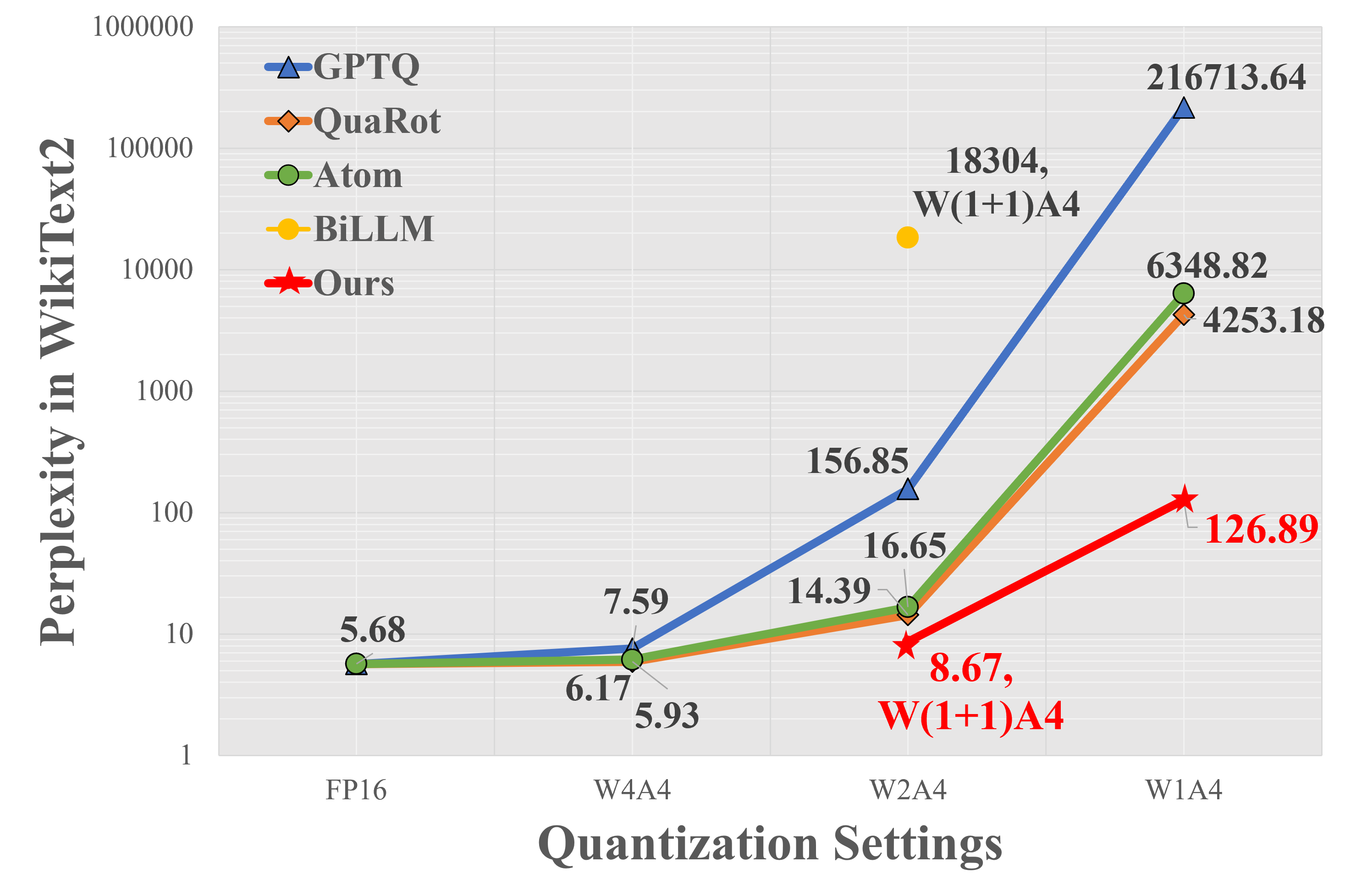}
\caption{The perplexity ($\downarrow$) of GPTQ, QuaRot, Atom, and our method on the LLAMA1-7B model under various quantization bit-width settings reveals that GPTQ, QuaRot, and Atom exhibit significant performance degradation when weights are quantized to 1 bit, whereas our method still demonstrates language generation capabilities close to those of the original model.}
\label{pic:overview}
\end{center}
\vskip -0.2in
\end{figure}

The enormous computational and memory overhead limits the widespread adoption of large-language models (LLMs). Model quantization methods \cite{nagel2021white, huang2024good} is a major approach to alleviate these challenges. Recent studies \cite{zhao2024atom}  on post-training quantization techniques (PTQ) can attain nearly lossless quantization under the W4A4 configuration.
However, lower-bit quantization of LLMs remains challenging. Existing methods \cite{zhao2024atom, ashkboos2024quarot} still suffer from the performance collapse issue when the number of bits for both weight and activation is less than 4, as shown in Figure \ref{pic:overview}.

Most of the PTQ LLM quantization work adopts the round-to-nearest (RTN) method \cite{nagel2021white} as the implementation scheme in practical operations, focusing primarily on how to preprocess the data for RTN quantization. For example,  LLM.int8 \cite{dettmers2022gpt3} and Atom \cite{zhao2024atom} handle outliers with high precision, effectively increasing the numerator of the scaling factor for some key values to reduce quantization error. Another line of works, including SmoothQuant \cite{xiao2023smoothquant}, QuaRot \cite{ashkboos2024quarot}, and DuQuant \cite{lin2024duquant}  smooth the distribution of data to be quantized via matrix rotation. Although these works can achieve satisfactory 4 bit quantization, they struggle to further reduce the quantization bit-width, which implies that the simple RTN quantization method is not sufficient  for very low-bit quantization with high performance. 


Another line of works adopt vector quantization technique \cite{liu2024vptq,van2024gptvq}. These approaches quantized the model weights via grouping the weights and assigning the weights in the same group to the same value. Those methods usually yield more accurate quantization models with smaller size, but cannot be speed-up the inference speed, as they have to dequantization the weight to recover float point format before computation.


To address these issues, we propose a quantization framework, that can simultaneously achieve small model size, fast computation, and high-quality inferece outputs. It consists of three parts: 1-bit weight quantization with another bit for fine-grain grouping, and $1\times 4 $ bits activation quantization using 4 times 1 bit channel to represent 4 bit quantization. By this way, we are able to compute the inner loop vector product in pure binary operations drastically boosting the inference speed as well as reduce the model size. 
%
%
The main contributions of this work are as follows:
\begin{itemize}
\item[$\bullet$] We propose  a  W(1+1)A(1×4) post-training quantization framework, 
with the boolean operations for the inner-loop multiplication-summation operations. 


\item[$\bullet$]  We present an EM-based algorithm for searching  boolean weight quantization with fine-grouping, noticeably improved the performance compared with the usual RTN approach. 


\item[$\bullet$] 
Our method  outperforms existing PTQ methods for 2 or 1 bit-widths. On the Wikitext2 benchmark, our method achieves perplexities of 8.58 and 8.89 on LLaMA-7B and LLaMA2-7B, respectively, using only the W2A4 quantization setting. This significantly surpasses the existing state-of-the-art methods and is  comparable to the performance of 5.68 and 5.47 achieved by the original full-precision models.
\end{itemize}



\section{Related Works}

Considerable research efforts have been devoted to the low-resource computation of LLMs \cite{dettmers2024qlora,chen2024recoverable}. Existing works have achieved near-lossless quantization at W4A4 through methods such as mixed-precision quantization \cite{zhao2024atom} and outlier smoothing using rotation factors \cite{ashkboos2024quarot, liu2024spinquant}. However, these methods encounter difficulties when attempting to push towards even lower-bit quantization below 4 bits. 

Recently, researchers start to seek  binarization methods \cite{courbariaux2016binarized, qin2022bibert} for LLMs. One line of works are based on  Quantization-Aware Training (QAT) method. For example, BitNet b1.58 \cite{wang2023bitnet}, BitNet a4.8 \cite{wang2024bitnet}, OneBit \cite{xu2024onebit}, and FBI-LLM \cite{ma2024fbi} have designed 1-bit Transformer architectures specifically for LLMs, replacing original linear layers  with specific quantized version. These methods can reduce the quantization bit-width of weights to 1 bit and demonstrate significant advantages over baseline models. However, these QAT methods require substantial computational resources, making it impractical for the quantization of LLMs.

Another line of works  explored the possibility of binarizing LLMs within the PTQ framework. BiLLM \cite{huang2024billm} and STBLLM \cite{dong2024stbllm}, inspired by the distribution characteristics of weight values and the Hessian matrix, adopt binary residual approximations for significant weights and optimal partitioning for group binarization of non-significant weights, which pushes the quantization boundary of LLM weights down to 1 bit. However,  these works neglect the quantization of activation values, which obstructs the computational acceleration of the quantized model and ultimately results in relatively slow inference speed. 

Most recently, people start to explore fully binarized models for both weight and activation. The works, ABQ-LLM \cite{zeng2024abq} and QBB \cite{bulat2024qbb}, propose the idea of decomposing a high-bit matrix into a set of binary matrices for accelerated computation. These methods hold promise for achieving INT1 computational acceleration but are constrained by the limitations of quantization bit-width and performance, preventing further improvements.

Compared with state-of-the-art (SOTA) W4A4 quantization efforts, our W(1+1)A(1×4) quantized model can further reduce computational and memory overhead, which further pushes the boundaries of LLM quantization and enriches the research on LLM binarization.

 \begin{figure}[tp]
\vskip 0.2in
\begin{center}
\includegraphics[width=6cm]{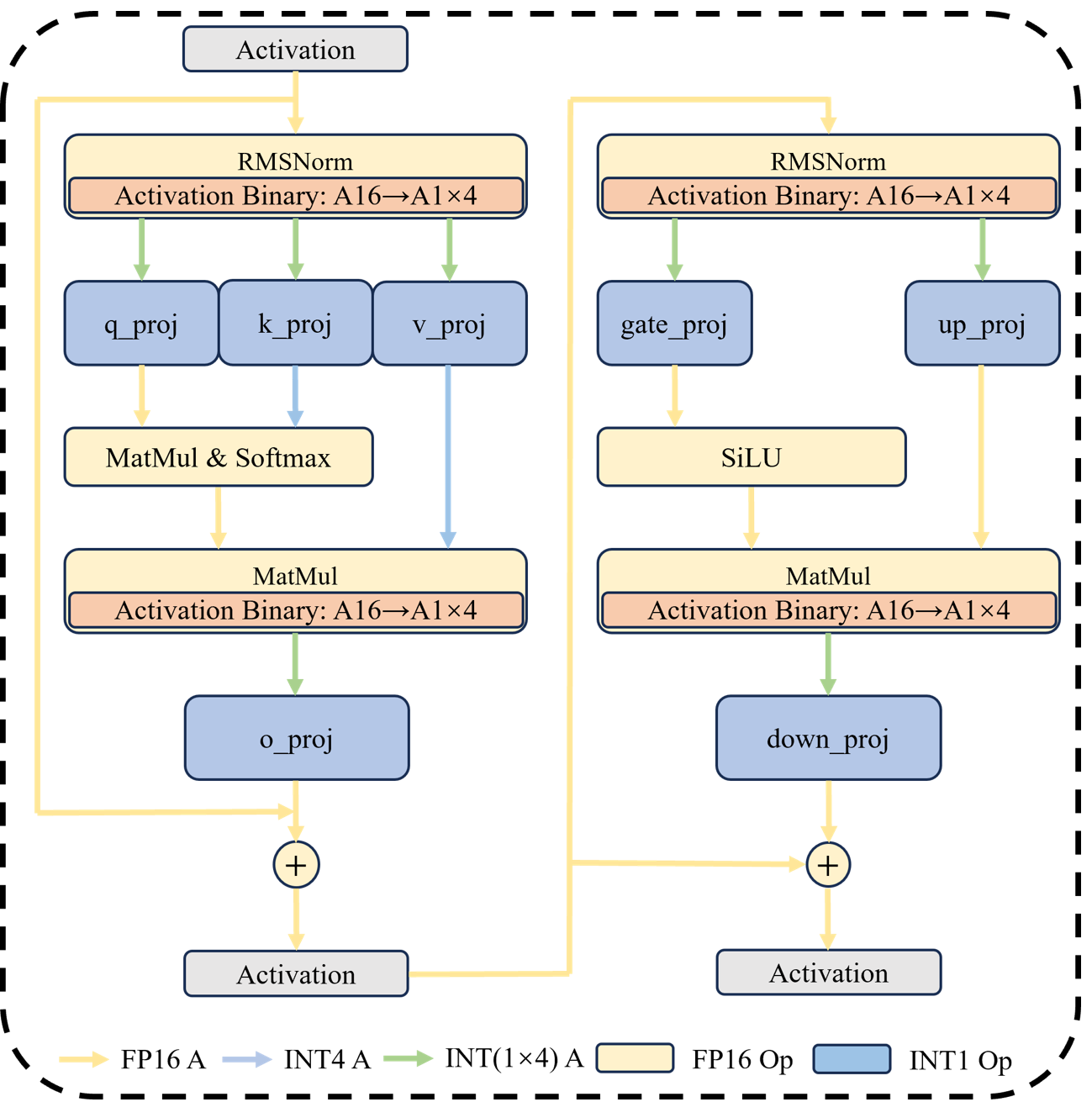}
\caption{
 binarized weight and activation  attention
\label{pic:method}
}
\end{center}
\vskip -0.2in
\end{figure}

\section{Method}
In this section, we introduce our binarized weight and activation (BWA) approach, a  paradigm designed to simultaneously accelerate inference speed and reduce memory footprint in large language models. We begin by outlining the architecture, and then describe an EM-based method for computing binarized weights with two-level grouping structure, and conclude by discussing the strategies in these work and  other relevant methods.


\subsection{Binarized weight and activation (BWA) attention}

%


\paragraph*{Overall structure of BWA attention} Our BWA  framework is  a quantization-aware modification of the standard attention module of a LLAMA-like model.

Following the dataflow of the activations  in Figure \ref{pic:method}, we introduce the overall structure. The input activation is a $C$-dimensional   FP16 vector. After the RMSNorm layer, we quantize the activation first from FP16 to INT4 using the standard Round-To-Nearest (RTN) method, then further transform it into four boolean variables. Next, the boolean activations are fed into a binarized fully-connected layers to compute key (K), query (Q), and value (V) matrices. The inner-loop computation is boolean, but the outputs recovers FP16 for query branch, and INT4 for key and value branches, {\em i.e.,} using 4 bit for KV cache. After the attention, we transform the activation vectors to boolean again so that all FC layer in the subsequent projection operations are also binarized. 


At the core of quantization task lies the binarization of fully connected (FC) layers, which account for approximately 90\% of memory bandwidth and computation. In contrast, other components such as normalization and attention score matrices contribute to the remaining 10\%. Consequently, our work primarily focuses on the binarization of FC layers, which are utilized in computing K, Q, V matrices, as well as all projection layers. Since the most computationally expensive operation in these layers is the matrix-vector multiplication within the inner loop, we pay most effort to design a computational efficiency FC layer, where the inner-loop only have binary operations.


\textbf{Binarized fully-connected layer } 
The core modification of the proposed BWA attention module is the binarization of the linear layer. 
To optimize the tradeoff between model accuracy and efficiency, we introduce a two-level weight grouping strategy for weight binarization, and a binarized decomposition to further quantize a 4 bit activations into four 1 bit boolean variables. Consequently, in the core inner-loop computation of the fully connected layers, only boolean operations are involved, drastically simplifying and boosting the computation. 


Specifically,  a FC layer computes a single token 
$
\boldsymbol{y}=\boldsymbol{W} \boldsymbol{x}
$, where $\boldsymbol{W} \in  \mathbb{R} ^ { {C_{\text{in}}} \times C_{\text{out}} }$, $\mat{x}$ is the $C_{\text{in}}$-dimensional  vector represented as an input token, and $\mat{y}$ is the output token vector.

\textbf{1) Channel-wise grouping   } 
Both weights and activations have a width dynamical range, which are hard to quantize using the same scaling parameter.
We sort the input channels by the average activation scales of test samples $\mat{X} $, {\em i.e., } sort according to $\text{diag}(\mat{X}\mat{X}^T)$ in an ascending order. Then, following the sorted order, we group the channels with similar activation scale. 
In our implementation, we divide the $C_{\text{out}}=4096$ channels into 32 groups. Each group has $B=128$ channels
\begin{equation} \label{eq:FC}
\begin{aligned}
y_j = \sum_{\ell=0}^{\lceil C_{\text{out}}/B \rceil-1 }\sum_{i=0}^{B-1} \boldsymbol{W}_{j,(B\ell +i)} \boldsymbol{x}_{B\ell + i}, \\j=1,2,\ldots,C_{\text{out}}.
\end{aligned}
\end{equation}
Note that channel-wise group is implemented   by a proper permutation of rows and columns of the weight matrix. Therefore, channel-wise grouping does not introduce any dditional computational or memory cost.

\textbf{2) Element-wise grouping} Directly binarization of $\mat{W}$ to 1 bit is not enough resulting severe model degeneration forbidding practical usage, especially combined the target to reduce the activation to 4 bit or lower.
Similar to BiLLM, we use additional element-level group to further group weight, and quantize them separately.

Define the set $D_{\ell,1} \subseteq \{0,1,2,\ldots,B-1\}$
, and its complementary $D_{\ell,0}= \{0,1,2,\ldots,B-1\} \setminus D_{\ell,1}$.
We quantized the two group separately using 1-bit for each weight element
\begin{equation} \label{eq:bW}
 \boldsymbol{W}_{j,(B\ell +i)} \approx \hat{\boldsymbol{W}}_{j,(B\ell +i)}  = \alpha_{j,\ell,s} q_{j,(B\ell +i)} + \beta_{j,\ell,s},
\end{equation}
for $s=1$ if $i\in D_\ell$ otherwise $s=0$, in which $\alpha_{j,\ell,s}$ and $\beta_{j,\ell,s}$ are two quantization parameters for each fine-grained group. 
In the standard LLAMA-7B model, each column-wise group contains has 128 channels. They are further categorized into two element-wise group, of which the sizes may not equal. The element-wise group are represented using a bit map. Together another bit $q_{j,i}$ representing weight sign. The weight matrix uses 1 + 1 bit per element. 

This fine-grained level grouping can effectively the accuracy at the expense of increase additional computational cost. On the other hand, with proper usage of bitmap operations, the cost can be suppressed to be marginal. The element-wise grouping has been used in the works \cite{huang2024billm, dong2024stbllm}, we improve this strategy by leveraging EM-based parameter searching with Hessian metric information to determine the optimal weighted split point in Section \ref{sec:3.2}.

\textbf{3) Activation binarization.}
The input activation  is quantized before feeding into the FC layer from FP16 to INT4 using the standard RTN method~\eqref{eq:RTN}. Specifically,  the quantization process of RTN to obtain $\boldsymbol{X}_q$ with $b$ bits is expressed as
\begin{equation}\label{eq:RTN}
    \begin{split}
    &\boldsymbol{X}_q =\text{clamp}(\lfloor \frac{\boldsymbol{X}}{\mu} \rceil+z,0,2^{k}-1),
    \end{split}
\end{equation}
%
where $\mu = \frac{\text{max}(\boldsymbol{X})-\text{min}(\boldsymbol{X})}{2^{k}-1}$ is the scaling parameter,$z = -\lfloor \frac{\text{min}(\boldsymbol{X})}{\mu} \rceil$ is the shift parameter. The clamp function denotes restricting the quantization result to the range between $[0,2^{k}-1]$. The notion $\lfloor\rceil$ signifies the nearest rounding operation.

To further binarize the activation, we first quantize the full-precision activation values $\boldsymbol{x}_i$ into 4 bit quantized values $x_{q,i}$, which can be further decomposed into four 1 bit Boolean variables $b_{i,a}$ with $a=0,1,2,3$. In the context of 4-bit quantization, $\hat{\boldsymbol{x}}_i$ represents the dequantized result
\begin{equation} \label{eq:bA}
\boldsymbol{x}_i \approx \hat{\boldsymbol{x}}_i = \mu (x_{q,i} + z) = \sum_{a=-1}^3\mu_{a} b_{i,a},
\end{equation}
where $\mu_{a} = \frac{2^a}{\sum_{a=0}^{3}2^a} \mu$ for $a=0,1,2,3$ and a special notation of  $\mu_{-1}=\mu z$ and $b_{i,-1}\equiv 1$ for the shift constant. Moreover, we can also relax $\mu_{k}$ as a free quantization parameter to be tuned manually or learn from data. To further enhance the performance of the quantized model, we adopted a scaling factor balancing strategy to minimize the L1 quantization error of activation values. Detailed implementation and analysis are provided in Appendix \ref{sec:binarizedresidualdecomposition}. Note that  as the weight is reordered and grouped, the elements of the input activation vector will be permuted accordingly. 

\textbf{4) Binarized FC layer}
We utilize the commutative property  of summation to make sure the inner multiplication summation is binary, so that the proposed method is actually boosting the computational speed. 

Substituting \eqref{eq:bW} and \eqref{eq:bA}
\begin{align} \nonumber
y_j
&\approx 
 \sum_{\ell=0}^{\lceil C_{\text{out}}/B \rceil-1 }\sum_{i=0}^{B-1} \Big[
 \big(\alpha_{j,\ell,s} q_{j,(B\ell +i)} + \beta_{j,\ell,s} \big) 
 \\ \nonumber
 &\quad\quad\quad\quad\quad\quad\quad
 \sum_{a=-1}^3\mu_{\ell,a} b_{(B\ell +i),a} \Big] 
\\ \nonumber
&=
\sum_{\ell=0}^{ \lceil C_{\text{in}}/B \rceil -1 }
 \sum_{a=-1}^{3}
 \mu_{\ell,a} 
 \sum_{s=0,1}
 \\
&\Big[ 
 \alpha_{j,\ell,s} v_{j,\ell,s,a} + \beta_{j,\ell,s}  r_{j,\ell,s,a}
\Big],
\end{align}
with the bit-wise inner product $v_{j,\ell,s,a} $ and  counting number of bits valued 1

\begin{equation}\label{eq:vr}
\begin{aligned}
&v_{j,\ell,s,a} =\sum_{i \in D_{j,\ell,s }} q_{j,(B\ell +i)} b_{(B\ell +i),a},\\& r_{j,\ell,s,a} =\sum_{i \in D_{j,\ell,s} }  b_{(B\ell +i),a}.
\end{aligned}
\end{equation}
Since the above two summation only involves bit variable $q_{j,(B\ell +i)}$, $ b_{(B\ell +i),a}$, and the set $D_{\ell,1}$ and $D_{\ell,2}$ is represented by a bitmap. Therefore,  they can be implemented efficiently by bit-wise XOR/AND and popc operation.
Specifically, let $e_{j,\ell,a} = q_{j,\ell} \land b_{\ell,a} \nonumber$ with $ q_{j,\ell}=[q_{j,B\ell},q_{j,(B\ell+1)},\ldots,q_{j,(B\ell+B-1)}]$ and $b_{\ell,a}=[b_{j,B\ell},b_{j,(B\ell+1)},\ldots,b_{j,(B\ell+B-1)}]$. Then, $v_{j,\ell,s,a}$, $r_{j,\ell,s,a}$ in  \eqref{eq:vr} can be computed by
\begin{equation}
\label{eq:bop}
\begin{aligned}
v_{j,\ell,s=0,a}&=\text{ Popc }( e_{j,\ell,a}  \land m_{j,\ell} ), \\
v_{j,\ell,s=1,a} &= \text{ Popc }( e_{j,\ell,a}  \land ( \lnot  m_{j,\ell} ) ), \\
r_{j,\ell,s=0,a}&=\text{ Popc }( b_{\ell,a}  \land m_{j,\ell} ), \\
 r_{j,\ell,s=1,a} &= \text{ Popc }( b_{\ell,a}  \land ( \lnot  m_{j,\ell} ), 
\end{aligned}
\end{equation}
where $m_{j,\ell}\in\{0,1\}^B$ is a bitmap indicting the fine-grain group affiliation with $[m_{j,\ell}]_i=0$ if $i \in D_{j,\ell,0}$, otherwise $m_i=1$.  

The inner loop, which originally be a multiplication-summation operation of 64 numbers, reduces to four bit-wise operations on 128 bits length variables, which can be efficiently implemented on both GPU and CPU.

%

\textbf{5) {Outlier activation } }
In this paper, we trick the last channel-wise group as outlier, and use 8 bit to quantize these channels. It is  hardware-friendly and can  efficiency implemented via reordering the channels \cite{yuan2023rptq, zhao2024atom}. Different from other methods, which usually use more channels for outliers, we only use minimal 1 group, which only contain 128 channels for the outliers (both in activation and weight). This minimizes the outlier overhead to approximately 3\% of the total channels in 8 bit and representing all other normal channels in 1 bit. Experiments for set more  outlier groups are done in Appendix \ref{sec:additionalexperiment}, which can further improve the performance at the expense of costing extra bits.



\subsection{Weight binarization and parameterization by Fine-Grained Group Hessian-Aware Quantization}\label{sec:3.2}
In this subsection, we propose an EM-based method to determine the binarized values of the weight matrices and the associated parameters for dequantization.

The weight matrix is quantized according to~\eqref{eq:bW}, where $q_{j,(B\ell+i)}$ represents the binarized weight, and $\alpha_{j,\ell,s}$, $\beta_{j,\ell,s}$ denote the scaling and shifting parameters, respectively, used for dequantization. Furthermore, we also require a bit map matrix of identical dimension to the weight matrix, which serves to determine the fine-group affiliation of each corresponding element.

We start from minimizing the L2 norm of the weight matrix $\mW$, utilizing an approximate Hessian weight as proposed in \cite{NIPS1992_303ed4c6, frantar2022optimal}
\begin{equation}
    \mathcal{L}_{\hat{\boldsymbol{W}}} = \Big\Vert \frac{1}{\text{diag}(\boldsymbol{H}^{-1})} (\boldsymbol{W}-\hat{\boldsymbol{W}}) \Big\Vert_{2}^2,
\end{equation}
where the Hessian matrix $\mH=\mX\mX^T$ is determined using a validation dataset, encapsulates information about the activation values during the computation of linear layers and gradient information during back propagation.

In our binarized parameterization \eqref{eq:bW}, each channel (indexed by $j$) and each channel-wise group (indexed by $\ell$) are parameterized independently.  Without loss of generality, we only focus on a single channel-wise group, which has $B$ weight elements $w_i$, $i\in{0,1,\ldots, B-1}$. 
The binary representation of $W$ with a binary element-wise group can at most have four different float-point values. Therefore, 
the quantization problem boils down to a 1-D clustering problem to determine 4 clusters, centered at $\hat{w}(0,0)$,  $\hat{w}(0,1)$,  $\hat{w}(1,0)$,  $\hat{w}(1,1)$. Formally, for each group, we solve the following minimization problem
\begin{equation} \label{con:em_objective}
\mathop{\min}_{\mat{s}, \mat{q} \in \{0,1 \}^{B} ,\hat{\mat{w}} \in {\mathbb{R}^4} } \sum_{i=0}^{B-1} \big(w_i - \hat{w}(s_i,q_i) \big)^2
/ {\text{diag}(\boldsymbol{H}^{-1})_{i}} \; ,
\end{equation}
where $\mat{s}$,  $\mat{q} $ represents  the fine-group affiliation and the binary weight value respectively.
Knowing the four values of $\hat{\mat{w} }$, one can recover the scaling and shifting parameters $\alpha_{j,\ell,s}$ and $\beta_{j,\ell,s}$ by the method of undetermined coefficients formalized as a set of 4-D linear equations.

We implement a revised EM algorithm to solve the above optimization problem, as shown in Algorithm \ref{alg:GHQframework}, The detailed steps are as follows:

\textbf{E-step:} Given the centroids $\hat{w}$, calculate the weighted distances from all weight elements $w$ in the current layer to each centroid, classify $w$ into the nearest class $C$, and determine the grouping accordingly, ensuring that the loss value in Equation (\ref{con:em_objective}) is minimized.

\textbf{M-step:} Given the grouping $D$ and the centroid class $C$ to which each weight element $w$ belongs, determine the $\mu$ centroid value that minimizes the loss in Equation (\ref{con:em_objective}) under the current grouping and centroid assignment.

In practical operation, the execution of the EM algorithm resembles a one-dimensional K-means clustering process. Additionally, to reduce the quantization error of the model, we incorporate the block compensation strategy from the GPTQ \cite{frantar2022gptq} framework into the quantization process. We conduct the update process in steps according to the block size set in GPTQ, with error compensation inserted between each step. This approach further reduces the overall quantization error of the model.

\begin{algorithm}[tb]
   \caption{Main Framework of our method}
   \label{alg:GHQframework}
\begin{algorithmic}[1]
   \REQUIRE ~~\\
   \quad $\textbf{W}\in\mathbb{R}^{C_{out}\times C_{in}}$, weight matrix\\
   \quad $\textbf{X}\in\mathbb{R}^{T\times C_{in}}$, calibration data\\
   \quad $B$, block size \\
   \quad $K$, outliers keep in INT8 \\
   \quad $iters$, EM steps
   \ENSURE ~~\\
   \quad $\textbf{B}$, weights after dequantization\\
   \quad $\textbf{D}$, fine-grained group information\\
   \STATE $\textbf{W} = \text{reorder}(\textbf{W}, \text{diag}(\textbf{X}\textbf{X}^{T}))$ 
   \STATE $\textbf{H} = 2\textbf{X}\textbf{X}^{T}$ \COMMENT{Hessian matrix}
   \STATE $\textbf{H}^{c}=\text{Cholesky}((\textbf{H}+\lambda\textbf{I})^{-1})$
   \STATE $\textbf{B} = 0_{C_{in}\times C_{out}}$
   \FOR{$i=0,B,2B,...,C_{out}-K-B$}
   \STATE $\textbf{W}^{p} = \textbf{W}_{:,i:i+B}$
   \STATE $\textbf{H}^{cp} = \text{diag}(\textbf{H}^{c}_{i:i+B,i:i+B})$
   \STATE $\textbf{C} = \text{init\_centers}(\textbf{W}^{p}, \textbf{H}^{cp})$
   \FOR{$j=1$ {\bfseries to} $iters$}
   \STATE $\textbf{D}=\text{get\_groups}(\textbf{W}^{p}, \textbf{H}^{cp}, \textbf{C})$ \COMMENT{E-step}
   \STATE $\textbf{I}=\text{get\_clusters}(\textbf{W}^{p}, \textbf{H}^{cp}, \textbf{C}, \textbf{D})$ \COMMENT{M-step}
   \STATE $\textbf{C} = \text{update\_centers}(\textbf{W}^{p}, \textbf{H}^{cp}, \textbf{I}, \textbf{D})$ \COMMENT{M-step}
   \ENDFOR
   \STATE $\textbf{B}_{:,i:i+B} = \text{binary}(\textbf{W}^{p}, \textbf{C}, \textbf{I}, \textbf{D})$ 
   \STATE $\textbf{E} = \frac{(\textbf{W}^{p} - \textbf{B})}{\textbf{H}^{cp}}$
   \STATE $\textbf{W}_{:,i+B:C_{out}-K} = \textbf{W}_{:,i+B:C_{out}-K} - \textbf{E}\cdot\textbf{H}^{c}_{i:i+B,i+B:}$
   \ENDFOR
   \STATE $\textbf{B}_{:,C_{out}-K:} = \text{quant\_int8}(\textbf{W}_{:,C_{out}-K:})$
   \STATE $\textbf{Return}\;\textbf{B}, \textbf{D}$
\end{algorithmic}
\end{algorithm}

\subsection{Remarks on quantization strategies}
We provide additional discussion on the quantization strategies employed in our approach, comparing them to existing methods such as the  RTN quantization, vector quantization for weight quantization, and a recent work on binarized residual decomposition for activation quantization.

\textbf{RTN quantization v.s. fine-grained group binary weight} 
In our method, we utilize 1 bit to store binary weights and an additional 1 bit to represent element-wise group affiliation, effectively using 2 bit of information. It enables each weight element to take on four distinct values. In contrast to the RTN quantization method widely used in other post-training quantization (PTQ) methods, where dequantized values are equally spaced, our model allows the four values to be chosen arbitrarily, which is optimized by the proposed EM-based algorithm.

\textbf{Vector quantization v.s. fine-group binary weight}
The optimization process for dequantization parameters in our approach is similar to that employed in vector-quantization-based methods \cite{frantar2022gptq, van2024gptvq}, where $2^n$ floating-point values are stored as representative values for $n$ bit quantization. In these approaches, dequantization must be performed before computing the vector inner product. In contrast, our method further parameterizes the representatives using binary weights and fine-grained group bits, along with floating-point scaling and shift parameters. This enables us to compute the vector inner product using pure Boolean operations as shown in \eqref{eq:bop}, resulting in a significant boost in computational speed. The details of quantization and dequantization after EM Algorithm are shown in Appendix \ref{sec:qanddq}.

\textbf{Binarized residual decomposition  and  $1\times 4$ bit representation of activation} 

The work \cite{zeng2024abq}  explored the approach of transforming arbitrary integer weight and activation WxAa into $xa \; \times$ W1A1 to achieve computational acceleration. It make use of bit operation to computation inner-loop vector product, but the original work can not get good below W4A4. On the other hand, the expansion of high bits weight and activation usually result heavy over head, as the number of channel vectors (relates to memory bandwidth) roughly from $(x+a)$ bits to $(xa)$ bits. In this work, we manage to reduce W(1+1)A(1×4), together with bitmap operation on the fine-grain group, the over-head cost is marginal.

\section{Experiments} 

\begin{table*}[h!]
\caption{Perplexity($\downarrow$) and Zero-shot QA accuracy($\uparrow$) results under the W4A4 and W2A4 settings on LLAMA1-7B and LLAMA2-7B. "FP16" denotes the performance of the original model represented in FLOAT16 format, with the best quantization performance highlighted in bold. The experimental results on the 13B model are presented in Table \ref{tab:13bppl}, and Table \ref{tab:13bacc}.}
\label{tab:llama7b}
\vskip 0.15in
\begin{tiny}
\begin{center}
\begin{tabular}{cclcccccccccc}
\toprule
\textbf{Model} & \textbf{Bits} & \textbf{Method} & \textbf{Wiki.}$\downarrow$ & \textbf{PTB}$\downarrow$ & \textbf{C4$\downarrow$} & \textbf{PIQA}$\uparrow$ & \textbf{ARC-E}$\uparrow$ & \textbf{ARC-C}$\uparrow$ & \textbf{BoolQ}$\uparrow$ & \textbf{Hella.}$\uparrow$ & \textbf{Wino.}$\uparrow$ & \textbf{Avg.}$\uparrow$ \\
\midrule
\multirow{7}{*}{\textbf{\makecell{LLAMA1\\-7B}}} & FP16 & - & 5.68 & 27.34 & 7.08 & 77.37 & 52.48 & 41.38 & 73.06 & 73.00 & 67.01 & 64.05 \\
\cline{2-13}
& \multirow{2}{*}{W4A4} & QuaRot & 6.41 & 49.73 & 8.43 & 74.81 & 50.13 & 38.74 & 70.98 &  68.80 & 61.56 & 61.01 \\
& & Atom & 6.30 & 30.28 & 7.98 & 75.35 & 51.60 & 36.69 & 70.86 & 67.27 & 64.33 & 62.21 \\
\cline{2-13}
& \multirow{2}{*}{W2A4} & QuaRot & 14.39 & 222.95 & 27.70 & 59.52 & 37.88 & 26.62 & 62.20 & 41.56 & 54.62 & 47.07 \\
& & Atom & 16.65 & 298.78 & 33.87 & 57.24 & 35.23 & 26.11 & 53.98 & 36.77 & 50.51 & 43.31 \\
\cline{2-13}
& W(1+1)A16 & BiLLM & 35.04 & 421.27 & 39.59 & 61.20 & 36.00 & 25.70 & 62.70 & 36.80 & 51.10 & 45.58 \\
\cline{2-13}
& W(1+1)A4 & BiLLM & 18304 & 17152 & 20736 & 50.05 & 25.38 & 26.54 & 49.63 & 26.05 & 49.49 & 37.86 \\
\cline{2-13}
 & W(1+1)A(1×4)&\cellcolor{lightgray!40}\textbf{Ours} &\cellcolor{lightgray!40} \textbf{8.58} &\cellcolor{lightgray!40} \textbf{76.09} &\cellcolor{lightgray!40} \textbf{12.27} &\cellcolor{lightgray!40} \textbf{68.88} &\cellcolor{lightgray!40} \textbf{45.03} &\cellcolor{lightgray!40} \textbf{30.89} &\cellcolor{lightgray!40} \textbf{69.63} &\cellcolor{lightgray!40} \textbf{55.41} &\cellcolor{lightgray!40} \textbf{59.35} &\cellcolor{lightgray!40} \textbf{54.87} \\
\hline
\multirow{7}{*}{\textbf{\makecell{LLAMA2\\-7B}}} & FP16 & - & 5.47 & 22.51 & 6.97 & 76.93 & 53.58 & 40.53 & 71.07 & 72.96 & 67.17 & 63.71 \\
\cline{2-13}
& \multirow{2}{*}{W4A4} & QuaRot & 6.32 & 71.21 & 8.67 & 74.32 & 51.60 & 38.23 & 68.41 & 69.24 & 61.56 & 60.89 \\
& & Atom & 6.18 & 27.94 & 8.05 & 75.24 & 52.74 & 37.12 & 71.16 & 67.89 & 63.93 & 62.58 \\
\cline{2-13}
& \multirow{2}{*}{W2A4} & QuaRot & 49.98 & 571.22 & 80.14 & 54.41 & 28.45 & 23.21 & 57.89 & 28.57 & 48.15 & 40.11 \\
& & Atom & 19.49 & 508.82 & 39.85 & 56.69 & 32.32 & 23.21 & 58.53 & 35.74 & 49.49 & 42.66 \\
\cline{2-13}
& W(1+1)A16 & BiLLM & 32.48 & 3877.38 & 40.52 & 60.60 & 36.20 & 24.40 & 61.80 & 34.80 & 52.40 & 45.03 \\
\cline{2-13}
& W(1+1)A4& BiLLM & 16128 & 17152 & 15168 & 50.22 & 26.30 & 27.90 & 45.23 & 26.10 & 49.88 & 37.61 \\
\cline{2-13}
 & W(1+1)A(1×4)&\cellcolor{lightgray!40}\textbf{Ours} &\cellcolor{lightgray!40} \textbf{8.89} &\cellcolor{lightgray!40} \textbf{69.46} &\cellcolor{lightgray!40} \textbf{12.74} &\cellcolor{lightgray!40} \textbf{68.72} &\cellcolor{lightgray!40} \textbf{46.13} &\cellcolor{lightgray!40} \textbf{30.55} &\cellcolor{lightgray!40} \textbf{66.12} &\cellcolor{lightgray!40} \textbf{55.76} &\cellcolor{lightgray!40} \textbf{58.01} &\cellcolor{lightgray!40} \textbf{54.22} \\
\bottomrule
\end{tabular}
\end{center}
\end{tiny}
\vskip -0.1in
\end{table*}

\begin{table*}[h!]
\caption{Perplexity($\downarrow$) and Zero-shot QA accuracy($\uparrow$) results under the W4A4 and W2A4 settings on Vicuna family. "FP16" denotes the performance of the original model represented in FLOAT16 format, with the best quantization performance highlighted in bold.}
\label{tab:vicuna}
\vskip 0.15in
\begin{tiny}
\begin{center}
\begin{tabular}{cclcccccccccc}
\toprule
\textbf{Model} & \textbf{Bits} & \textbf{Method} & \textbf{Wiki.}$\downarrow$ & \textbf{PTB}$\downarrow$ & \textbf{C4$\downarrow$} & \textbf{PIQA}$\uparrow$ & \textbf{ARC-E}$\uparrow$ & \textbf{ARC-C}$\uparrow$ & \textbf{BoolQ}$\uparrow$ & \textbf{Hella.}$\uparrow$ & \textbf{Wino.}$\uparrow$ & \textbf{Avg.}$\uparrow$ \\
\midrule
\multirow{6}{*}{\textbf{\makecell{Vicuna\\-v1.5-7B}}} & FP16 & - & 6.78 & 26.78 & 8.55 & 77.80 & 56.06 & 39.93 & 75.69 & 71.06 & 67.80 & 64.72 \\
\cline{2-13}
& \multirow{2}{*}{W4A4} & QuaRot & 7.80 & 52.44 & 10.87 & 73.67 & 53.20 & 37.71 & 72.45 & 67.66 & 60.93 & 62.12 \\
& & Atom & 7.22 & 31.75 & 9.36 & 75.14 & 55.60 & 37.63 & 77.25 & 67.08 & 64.40 & 64.42 \\
\cline{2-13}
& \multirow{2}{*}{W2A4} & QuaRot & 39.51 & 226.50 & 65.17 & 55.66 & 33.38 & 22.75 & 62.08 & 31.71 & 50.51 & 44.03 \\
& & Atom & 15.96 & 107.68 & 25.13 & 56.64 & 31.90 & 29.61 & 64.07 & 46.30 & 55.33 & 47.31 \\
\cline{2-13}
 & W(1+1)A(1×4)&\cellcolor{lightgray!40}\textbf{Ours} &\cellcolor{lightgray!40} \textbf{9.51} &\cellcolor{lightgray!40} \textbf{45.61} &\cellcolor{lightgray!40} \textbf{13.35} &\cellcolor{lightgray!40} \textbf{69.97} &\cellcolor{lightgray!40} \textbf{50.00} &\cellcolor{lightgray!40} \textbf{33.45} &\cellcolor{lightgray!40} \textbf{71.96} &\cellcolor{lightgray!40} \textbf{57.81} &\cellcolor{lightgray!40} \textbf{59.43} &\cellcolor{lightgray!40} \textbf{57.10} \\
\hline
\multirow{6}{*}{\textbf{\makecell{Vicuna\\-v1.5-13B}}} & FP16 & - & 5.95 & 25.15 & 7.78 & 78.40 & 56.44 & 44.80 & 76.51 & 74.63 & 69.06 & 66.64 \\
\cline{2-13}
& \multirow{2}{*}{W4A4} & QuaRot & 6.81 & 54.16 & 9.64 & 74.81 & 51.43 & 40.53 & 70.73 & 70.96 & 62.12 & 62.51 \\
& & Atom & 6.32 & 27.64 & 8.25 & 76.44 & 54.67 & 43.34 & 74.83 & 72.07 & 66.46 & 65.37 \\
\cline{2-13}
& \multirow{2}{*}{W2A4} & QuaRot & 18.32 & 273.86 & 37.86 & 56.69 & 36.49 & 26.11 & 62.42 & 38.40 & 53.59 & 45.54 \\
& & Atom & 19.84 & 174.63 & 36.39 & 54.95 & 34.13 & 25.68 & 61.74 & 37.55 & 52.17 & 44.37 \\
\cline{2-13}
 & W(1+1)A(1×4)&\cellcolor{lightgray!40}\textbf{Ours} &\cellcolor{lightgray!40} \textbf{7.91} &\cellcolor{lightgray!40} \textbf{49.71} &\cellcolor{lightgray!40} \textbf{11.45} &\cellcolor{lightgray!40} \textbf{71.44} &\cellcolor{lightgray!40} \textbf{52.36} &\cellcolor{lightgray!40} \textbf{38.65} &\cellcolor{lightgray!40} \textbf{68.93} &\cellcolor{lightgray!40} \textbf{62.34} &\cellcolor{lightgray!40} \textbf{62.35} &\cellcolor{lightgray!40} \textbf{59.35} \\
\bottomrule
\end{tabular}
\end{center}
\end{tiny}
\vskip -0.1in
\end{table*}

\begin{table*}[h!]
\caption{Massive Multitask Language Understanding (MMLU) results (\%) under the W2A4 settings on LLAMA1-7B. "FP16" denotes the performance of the original model represented in FLOAT16 format, with the best quantization performance highlighted in bold.}
\label{tab:mmlu}
\vskip 0.15in
\begin{scriptsize}
\begin{center}
\begin{tabular}{cclccccc}
\toprule
\textbf{Model} & \textbf{Bits} & \textbf{Method} & \textbf{MMLU-STEM}$\uparrow$ & \textbf{MMLU-humanities}$\uparrow$ & \textbf{MMLU-social science}$\uparrow$ & \textbf{MMLU-others}$\uparrow$ & \textbf{Avg.}$\uparrow$ \\
\midrule
\multirow{3}{*}{\textbf{\makecell{LLAMA1\\-7B}}} & FP16 & - & 30.9 & 33.2 & 37.9 & 37.9 & 34.8 \\
\cline{2-8}
& W2A4 & Atom & 25.6 & 24.0 & 23.6 & 27.6 & 25.1 \\
\cline{2-8}
 & W(1+1)A(1×4)&\cellcolor{lightgray!40}\textbf{Ours} &\cellcolor{lightgray!40} \textbf{29.4} &\cellcolor{lightgray!40} \textbf{26.0} &\cellcolor{lightgray!40} \textbf{30.5} &\cellcolor{lightgray!40} \textbf{27.2} &\cellcolor{lightgray!40} \textbf{28.0} \\
\bottomrule
\end{tabular}
\end{center}
\end{scriptsize}
\vskip -0.1in
\end{table*}

\textbf{Setup.} We implemented our method on the PyTorch \cite{paszke2019pytorch} framework, where all linear layer weights in the original model are quantized to 1+1 bit, and input activations of all linear layers are quantized to 1×4 bits. For weights, we adopt per-channel asymmetric quantization with a clipping ratio set to 1.0 across all experiments, utilizing the GPTQ quantization framework to compensate for quantization errors. For activations, we employ per-token asymmetric quantization with a clipping ratio of 1.0. To optimize performance, we use RTN for dynamic quantization of the activation matrix. For KV caches, we uniformly apply 4 bits quantization to store and load. The quantization group size is 128, and the number of outlier channels is 128 (approximately 3\% of all channels). We use 128 random samples from the WikiText2 \cite{merity2016pointer} training set as the calibration dataset, with a sequence length of 2048. All experiments were conducted more than three times, and the average values were recorded.

\textbf{Models and Datasets.} We apply our method to the open-source LLAMA1 (7B, 13B) \cite{touvron2023llama}, LLAMA2 (7B, 13B) \cite{touvron2023llama2}, and Vicuna (7B, 13B) \cite{chiang2023vicuna} models and evaluate their performance on language generation, commonsense QA, and language understanding tasks. The primary metric for language generation tasks is perplexity, assessed on datasets including WikiText2, PTB \cite{marcus1994penn}, and C4 \cite{raffel2020exploring}. For commonsense QA tasks, the main metric is zero-shot accuracy, evaluated on datasets such as PIQA \cite{bisk2020piqa}, ARC \cite{clark2018think}, BoolQ \cite{clark2019boolq}, HellaSwag \cite{zellers2019hellaswag}, and WinoGrande \cite{sakaguchi2021winogrande}. Except for the C4 dataset, where we randomly select 256 samples of length 2048 from the test set for evaluation, we utilize the entire test set portion of these datasets for our testing. For language understanding tasks, we mainly tested our method on the Massive Multitask Language Understanding (MMLU) \cite{hendrycks2020measuring} benchmark.

\textbf{Baseline.} We compare our approach with state-of-the-art (SOTA) PTQ methods for weights and activations. Since few existing methods explore the W2A4 quantization setting, we implement W2A4 quantization for all compared methods to ensure fairness before evaluation. Our main baselines include Atom \cite{zhao2024atom}, QuaRot \cite{ashkboos2024quarot}, and BiLLM \cite{huang2024billm}. Atom and QuaRot are SOTA methods under the W4A4 quantization setting, while BiLLM is the SOTA for the W(1+1)A16 quantization setting.

\subsection{Main Results}

\textbf{Language Generation Tasks.} We assess the perplexity of our method on language generation tasks and conduct a fair comparison with existing SOTA methods. As shown in Table \ref{tab:llama7b} and Table \ref{tab:vicuna}, Atom and QuaRot, as SOTA methods under the W4A4 setting, experience significant performance drops under the W2A4 setting. In contrast, our method significantly outperforms these methods on all datasets under the W(1+1)A(1×4) setting which is equivalent to W2A4, and our method's perplexity evan approaches that of the FP16 model. It is noteworthy that BiLLM also utilizes an additional 1 bit to store extra fine-grained grouping information, thus we consider it as a W(1+1)A16 approach. As a similar method that employs fine-grained grouping like our method, its performance under the W(1+1)A16 configuration is significantly outperformed by our method with the W(1+1)A(1×4) setting. Furthermore, when its activation values are quantized to 4 bits, the performance of BiLLM rapidly deteriorates.


\textbf{Zero-Shot Tasks.} We also evaluate our method on six important zero-shot tasks. Table \ref{tab:llama7b} and Table \ref{tab:vicuna} presents the comparison results between our method and the baselines. our method significantly outperforms existing methods under the W2A4 quantization setting and demonstrates stable accuracy, approaching the performance of the FP16 model.

\textbf{Language Understanding Tasks.} We simultaneously evaluated the language understanding performance of our quantization method across 47 distinct subdomains of downstream tasks in the MMLU benchmark. Categorized performance reports in Table \ref{tab:mmlu} demonstrate that our approach surpasses existing SOTA quantization solutions while approaching the capabilities of the full-precision LLAMA1-7B model.

\subsection{Performance Analysis}

\begin{figure}[t]
\vskip 0.2in
\begin{center}
\centerline{\includegraphics[width=\columnwidth]{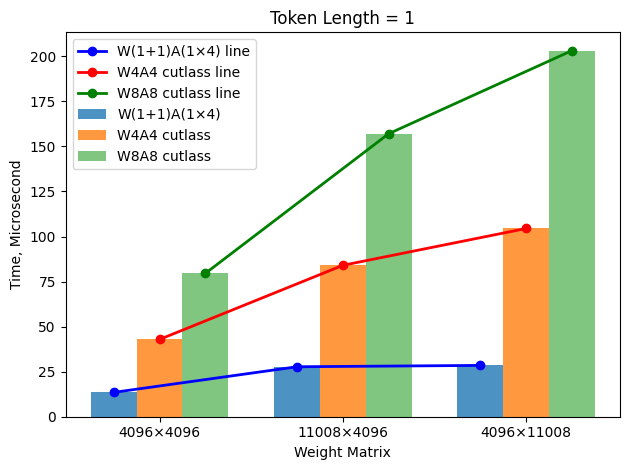}}
\caption{A comparison of the time cost between the W(1+1)A(1×4) kernel and the INT4, INT8 kernels of CUTLASS for matrix multiplication on the A6000. More results are presented in Figure \ref{pic:speed_full}.}
\label{pic:times}
\end{center}
\vskip -0.2in
\end{figure}

\begin{table*}[htb]
\caption{Ablation experiments on the effects of Minimum distance quantization (EM) and fine-grained group, with all experimental results based on LLAMA1-7B and a group size of 128.}
\label{tab:ablationofemandgroup}
\vskip 0.15in
\begin{center}
\begin{footnotesize}
\begin{tabular}{cc|cc}
\toprule
\textbf{Minimum distance quantization}& \textbf{Fine-grained group} & \textbf{Wiki. PPL}$\downarrow$ & \textbf{Avg. Accuracy}$\uparrow$\\
\midrule
\XSolidBrush & \XSolidBrush & 6348.82 & 36.44 \\
\checkmark & \XSolidBrush & 126.89 & 36.00 \\
\XSolidBrush & \checkmark & 16.65 & 43.31 \\
\checkmark & \checkmark & 8.58 & 54.87 \\
\bottomrule
\end{tabular}
\end{footnotesize}
\end{center}
\vskip -0.1in
\end{table*}

\begin{table}[htb]
\caption{Ablation experiments on the effects of different quantized components used in our method, with all experimental results based on LLAMA1-7B and a group size of 128.}
\label{tab:ablation}
\vskip 0.15in
\begin{center}
\begin{footnotesize}
\begin{tabular}{ll}
\toprule
\textbf{Quantization Method}& \textbf{Wiki.} $\downarrow$ \\
\midrule
LLAMA-7B FP16    & 5.68 \\
\hline
W1A4 GPTQ (Group size 128) & 216713 \\
+ Keep 128 outlier channels in INT8 & 6749 \\
+ Minimum distance quantization  & 126.89 \\
+ Fine-grained group, W(1+1)   & 8.69 \\
+ Hessian-weighted distance metric    & 8.65 \\
+ Binarized Residual Decomposition, A(1×4)    & 8.58 \\
\bottomrule
\end{tabular}
\end{footnotesize}
\end{center}
\vskip -0.1in
\end{table}

\textbf{Speedup.} To evaluate the inference acceleration provided by our method, we adopt the kernel implementation from ABQ-LLM \cite{zeng2024abq}, which supports decomposing arbitrary-dimensional WxAx operations into multiple W1A1 computations and leverages the acceleration effect of INT1 multiplication for significant speedup in matrix multiplication. We test the speedup of our method under the W(1+1)A(1×4) setting on an A6000 compared to different bit-width settings supported by CUTLASS, such as W8A8 and W4A4. Because the weights actually involved in the computations in our method are 1 bit, with an additional 1 bit solely used for storing fine-grained grouping information, we consider our method as a quantization method that can be viewed as a decomposition from W2A4 downwards in terms of computational acceleration comparison. As shown in Figure \ref{pic:times}, in terms of single-layer matrix computations, our method exhibits a more substantial speedup in comparison to other bit-width settings, surpassing the kernel acceleration of CUTLASS by a factor of 3 in matrix computations. This demonstrates that the approach of using INT1 for acceleration in our method can fully leverage the speedup benefits of low-bit computations. Moreover, since both the weights and activations in our method are quantized to very low bit-widths, the additional computational overhead introduced by the decomposition does not significantly impact the gains achieved through INT1 computation.

\subsection{Ablation Studies}

To evaluate the effectiveness of different quantization modules in our method, we compared the accuracy gains or losses among various quantization techniques employed within our method. The results presented in Table \ref{tab:ablation} demonstrate that outlier handling, Minimum distance quantization, and fine-grained grouping, as the basic processing schemes in our method. Each step significantly enhances the performance of the quantized model, effectively mitigating the performance collapse issue observed in the W1A4 quantized model. The introduction of the Hessian-weighted distance metric and binarized residual decomposition further boosts the quantized model's performance. Although numerically, the improvements in perplexity brought about by these two methods are not substantial, this is because the previous enhancements have already pushed the performance metrics close to those of the original model, leaving limited room for further improvement. Theoretically, the Hessian-weighted distance metric reveals a measure of weight importance, while the binarized residual decomposition elucidates the direction of performance enhancement after binarization decomposition.

\section{Limitations}

Although our our method can achieve fully binarized computation to reduce computational overhead during inference, a trade-off must be made regarding the accuracy of the quantized model. Consequently, we need to utilize additional bits to store the information of the quantized matrices, with the actual storage bits for weights and activations equivalent to 2 bits and 4 bits, respectively. This implies that our model has not been compressed to the theoretical extreme of a boolean model, leaving room for further improvement. Meanwhile, although the performance of our our method across various evaluation tasks is close to that of the pre-quantized model, it is not truly lossless quantization. This loss indicates that the quantized model has not fully restored the representational capacity of the original model. In the future, we consider employing methods that integrate quantization-aware training to further enhance the efficiency and performance of our method.

\section{Acknowledgement}

This work was supported in part by National Science and Technology Major Project under Grant 2022ZD0116500, National Natural Science Foundation of China under Grant 62222609, CAS Project for Young Scientists in Basic Research under Grant YSBR-083, Pioneer Hundred Talents Program of CAS under Grant Y9S9MS08 and Grant E2S40101, and Beijing Municipal Science and Technology Project under Grant No. Z231100010323005.
\bibliography{custom}

\appendix

\section{Binarized Residual Decomposition} \label{sec:binarizedresidualdecomposition}
Previous work \cite{zeng2024abq} has explored the approach of transforming arbitrary WxYx into $xy \; \times$ W1A1 to achieve computational acceleration. Building upon this foundation, we further investigate the use of residual smoothing scaling factors to enhance the quantization performance of LLMs.

\textbf{Binarized Decomposition} Specifically, we initially demonstrate that any matrix represented in the INT format can be decomposed into a combination of multiple INT1 matrices. The advantage of this approach lies in leveraging the efficient computation of W1A1 to replace the original high-bit computations. Obviously, replacing the original matrix multiplication with INT1 computations will result in a multiple increase in computational load, which is particularly pronounced when the original matrix computations involve high bits. However, our low-bit W1A4 quantization setting only increases the computational load to 4 times the original, which is acceptable given the computational speedup brought by the INT1 operators.

\textbf{Scaling factor balancing} Taking symmetric quantization as an example, the dequantization process of the activation values after binary decomposition is denoted as

\begin{equation}
    \begin{split}
    \textbf{X}_{\text{DeQuant}}=\mu\textbf{X}_{\text{INT4}} = \underbrace{2^3\cdot\mu}_{\mu_3} \cdot \textbf{B}_3 + \underbrace{2^2\cdot\mu}_{\mu_2} \cdot \textbf{B}_2 \\+ \underbrace{2^1\cdot\mu}_{\mu_1} \cdot \textbf{B}_1 + \underbrace{2^0\cdot\mu}_{\mu_0} \cdot \textbf{B}_0,
    \end{split}
\end{equation}

where $\textbf{X}_{\text{DeQuant}}$ represents the result of dequantization, and $\mu$ signifies the scaling factor within the quantization parameters. Unlike the conventional 4 bit dequantization process, where only one scaling factor is adjustable, we have the flexibility to independently alter the scaling factors $\mu_0$, $\mu_1$, $\mu_2$, and $\mu_3$ for each of the four INT1 matrices, under the premise that the quantized INT4 matrix can be equivalently replaced by a 4 $\times$ INT1 matrix. This adjustment aims to further reduce the quantization error of the activation values. Given that the initial 4 bit quantization demonstrates relatively good performance, we leverage these quantization parameters as prior knowledge to calculate the relative error $\mathbb{E}$ between the dequantized matrix and the original matrix. This error is then smoothly distributed across $\mu_0$, $\mu_1$, $\mu_2$, and $\mu_3$

\begin{equation}
\begin{aligned}
    \begin{split}
    \mu_i = \mu_i + \text{Avg}(\frac{\mu_i\textbf{B}_i}{\mu\textbf{X}_{\text{INT4}}}\odot\mathbb{E}),\\
    \mathbb{E} = \textbf{X}_{\text{FP16}} - \textbf{X}_{\text{DeQuant}}.
    \label{equ:smoothscalefator}
    \end{split}
\end{aligned}
\end{equation}

We use $\odot$ to denote element-wise multiplication of matrices and a fraction to represent element-wise division. $\textbf{X}_{\text{FP16}}$ stands for the original value matrix of the activations before quantization. The value of $i$ ranges from 0 to 3. Based on the Equation \ref{equ:smoothscalefator}, we minimize the first-order overall quantization error $\mathbb{E}$ to zero while preserving the distribution of quantized values in the original low-bit setting, thereby further enhancing the performance of the quantized model.

\section{Quantization and Dequantization after EM Algorithm} \label{sec:qanddq}
\textbf{Quantization and Dequantization} The quantization results obtained by EM Algorithm are akin to those of vector quantization \cite{frantar2022gptq, van2024gptvq}, in that it categorizes the original weight elements into $2^k$ values, rather than obtaining a low-bit INT quantized representation and quantization parameters as in RTN. In reality, this approach in vector quantization may be suboptimal because the $2^k$ values are not equally spaced and cannot be converted into a pure low-bit INT representation, thus precluding the quantized model from utilizing efficient low-bit matrix multiplication to accelerate inference. However, in our weight binarization process, the clustering centers (totaling $2^1=2$) obtained through the EM algorithm consistently maintain equal spacing and can be transformed into an INT1 representation. The transformation process can be expressed as

\begin{equation}
    \begin{split}
    \hat{\textbf{W}}_{i,j} &= \alpha\cdot\textbf{Q}(\textbf{W}_{i,j}) + \beta
    , \\ \textbf{Q}(\textbf{W}_{i,j}) &= \left\{
    \begin{aligned}
    1, \quad  \textbf{W}_{i,j} \in C_1\\
    -1, \quad  \textbf{W}_{i,j} \in C_2\\
    \end{aligned}
    \right. .
    \end{split}
\end{equation}

Here, The quantization results obtained by EM Algorithm are $C = \{C_1, C_2\}, \; \mu = \{\mu_1, \mu_2\}$, from which the dequantization parameters can be calculated as $\alpha=\frac{\mu_1-\mu_2}{2}, \; \beta = \frac{\mu_1+\mu_2}{2}$. Furthermore, the results of EM Algorithm are transformed into a form that enables efficient INT1 computation.

\section{Additional Experimental Results} \label{sec:additionalexperiment}

\begin{table}[h!]
\caption{Model size comparison of LLAMA family.}
\label{tab:gpumem}
\vskip 0.15in
\begin{center}
\begin{small}
\begin{tabular}{lcc}
\toprule
\textbf{Models}& \textbf{FP16} & \textbf{Ours} \\
\midrule
\textbf{LLAMA-7B} & 13.5GB & 2.69GB \\
\textbf{LLAMA-13B} & 24.2GB & 4.82GB \\
\textbf{LLAMA-30B} & 60.5GB & 12.05GB \\
\textbf{LLAMA-65B} & 121.0GB & 24.11GB \\
\bottomrule
\end{tabular}
\end{small}
\end{center}
\vskip -0.1in
\end{table}

\textbf{Time required for quantization.} Our method quantizes the weight matrices within all linear layers of the full-precision model. The quantization process for the 7B model can be completed in approximately 20 minutes, while the 13B model requires only about 30 minutes.

\smallskip\noindent\textbf{Results on 13B models.} As shown in Table \ref{tab:13bppl} and Table \ref{tab:13bacc}, we evaluated the performance of our method and other quantization methods on language generation tasks and zero-shot QA task accuracy using LLAMA1-13B and LLAMA2-13B models. Our findings indicate that, in general, the model performance adheres to the principle that increasing the number of model parameters leads to improved model performance. Furthermore, our method achieved state-of-the-art results across all evaluated metrics.

\begin{table*}[h!]
\caption{Perplexity($\downarrow$) results under the W4A4 and W2A4 settings on LLAMA1-13B and LLAMA2-13B. "FP16" denotes the performance of the original model represented in FLOAT16 format, with the best quantization performance highlighted in bold.}
\label{tab:13bppl}
\vskip 0.15in
\begin{scriptsize}
\begin{center}
\begin{tabular}{cclrrrcclrrr}
\toprule
\multirow{2}{*}{\textbf{Model}} & \multirow{2}{*}{\textbf{Bits}} & \multirow{2}{*}{\textbf{Method}} & \multicolumn{3}{c}{\textbf{Perplexity}$\downarrow$} & \multirow{2}{*}{\textbf{Model}} & \multirow{2}{*}{\textbf{Bits}} & \multirow{2}{*}{\textbf{Method}} & \multicolumn{3}{c}{\textbf{Perplexity}$\downarrow$}\\
\cline{4-6} \cline{10-12} 
 & & & \textbf{Wiki.} & \textbf{PTB} & \textbf{C4} & & & & \textbf{Wiki.} & \textbf{PTB} & \textbf{C4}\\
\midrule
\multirow{6}{*}{\textbf{\makecell{LLAMA1\\-13B}}} & FP16 & - & 5.09 & 19.23 & 6.61 & \multirow{6}{*}{\textbf{\makecell{LLAMA2\\-13B}}} & FP16 & - & 4.88 & 28.87 & 6.47\\
\cline{2-6} \cline{8-12}
& \multirow{2}{*}{W4A4} & QuaRot      & 5.71 & 36.10 & 7.57 & & \multirow{2}{*}{W4A4} & QuaRot      & 5.59 & 64.27 & 7.84\\
&  & Atom                             & 5.47 & 22.16 & 7.04 & &  & Atom                             & 5.26 & 32.46 & 6.95\\
\cline{2-6} \cline{8-12}
& \multirow{2}{*}{W2A4} & QuaRot      & 11.14 & 156.30 & 20.80 & & \multirow{2}{*}{W2A4} & QuaRot      & 17.49 & 386.40 & 38.88\\
&  & Atom                             & 11.69 & 115.62 & 19.55 & &  & Atom                             & 11.24 & 152.68 & 18.15\\
\cline{2-6} \cline{8-12}
 & W(1+1)A(1×4) &\cellcolor{lightgray!40}\textbf{Ours}                    &\cellcolor{lightgray!40} \textbf{7.19} &\cellcolor{lightgray!40} \textbf{37.20} &\cellcolor{lightgray!40} \textbf{10.18} & & W(1+1)A(1×4)&\cellcolor{lightgray!40}\textbf{Ours}                    &\cellcolor{lightgray!40} \textbf{7.17} &\cellcolor{lightgray!40} \textbf{56.91} &\cellcolor{lightgray!40} \textbf{10.44} \\

\bottomrule
\end{tabular}
\end{center}
\end{scriptsize}
\vskip -0.1in
\end{table*}

\begin{table*}[h!]
\caption{Zero-shot QA accuracy($\uparrow$) results under the W4A4 and W2A4 settings on LLAMA1-13B and LLAMA2-13B. "FP16" denotes the performance of the original model represented in FLOAT16 format, with the best quantization performance highlighted in bold.}
\label{tab:13bacc}
\vskip 0.15in
\begin{scriptsize}
\begin{center}
\begin{tabular}{cclccccccc}
\toprule
\multirow{2}{*}{\textbf{Model}} & \multirow{2}{*}{\textbf{Bits}} & \multirow{2}{*}{\textbf{Method}} & \multicolumn{7}{c}{\textbf{Zero-shot Accuracy}$\uparrow$} \\
\cline{4-10}
 & & & \textbf{PIQA} & \textbf{ARC-E} & \textbf{ARC-C} & \textbf{BoolQ} & \textbf{HellaSwag} & \textbf{WinoGrande} & \textbf{Avg.}\\
\midrule
\multirow{6}{*}{\textbf{LLAMA1-13B}} & FP16 & - & 79.05 & 59.89 & 44.71 & 68.47 & 76.23 & 70.24 & 66.43 \\
\cline{2-10}
& \multirow{2}{*}{W4A4} & QuaRot      & 76.61 & 55.30 & 41.64 & 67.09 & 73.02 & 65.59 & 64.24 \\
&  & Atom                             & 77.64 & 58.38 & 41.81 & 68.50 & 73.75 & 65.98 & 64.64 \\
\cline{2-10}
& \multirow{2}{*}{W2A4} & QuaRot      & 64.42 & 41.50 & 28.75 & 63.36 & 48.49 & 56.67 & 50.53 \\
&  & Atom                             & 59.68 & 36.62 & 29.01 & 58.56 & 44.84 & 52.01 & 46.79 \\
\cline{2-10}
 &W(1+1)A(1×4) &\cellcolor{lightgray!40}\textbf{Ours}                    &\cellcolor{lightgray!40} \textbf{72.09} &\cellcolor{lightgray!40} \textbf{48.57} &\cellcolor{lightgray!40} \textbf{34.13} &\cellcolor{lightgray!40} \textbf{62.54} &\cellcolor{lightgray!40} \textbf{62.63} &\cellcolor{lightgray!40} \textbf{64.88} &\cellcolor{lightgray!40} \textbf{57.47} \\
\hline
\multirow{6}{*}{\textbf{LLAMA2-13B}} & FP16 & - & 79.00 & 57.95 & 44.28 & 69.02 & 76.58 & 69.69 & 66.09 \\
\cline{2-10}
& \multirow{2}{*}{W4A4} & QuaRot      & 76.93 & 52.10 & 40.70 & 68.10 & 72.70 & 62.51 & 62.99 \\
&  & Atom                             & 77.37 & 56.73 & 42.32 & 67.62 & 74.07 & 68.27 & 65.46 \\
\cline{2-10}
& \multirow{2}{*}{W2A4} & QuaRot      & 59.09 & 34.60 & 24.23 & 62.11 & 35.03 & 51.38 & 44.41 \\
&  & Atom                             & 61.15 & 40.36 & 29.52 & 61.56 & 45.62 & 51.22 & 48.24 \\
\cline{2-10}
 &W(1+1)A(1×4) &\cellcolor{lightgray!40}\textbf{Ours}                    &\cellcolor{lightgray!40} \textbf{71.98} &\cellcolor{lightgray!40} \textbf{49.92} &\cellcolor{lightgray!40} \textbf{36.26} &\cellcolor{lightgray!40} \textbf{65.90} &\cellcolor{lightgray!40} \textbf{60.52} &\cellcolor{lightgray!40} \textbf{61.80} &\cellcolor{lightgray!40} \textbf{57.73} \\
\bottomrule
\end{tabular}
\end{center}
\end{scriptsize}
\vskip -0.1in
\end{table*}

\smallskip\noindent\textbf{Results of different outlier channel number settings.} In Table \ref{tab:outlierchannels}, we compare the relationship between different numbers of outlier channels and the quantization performance of our method. Since the group size is set to 128, we also use 128 as the unit here. The results demonstrate that preserving a small number of outliers with high precision can ensure overall quantization performance. Furthermore, when the number of outlier channels is increased, the model performance exhibits a nearly linear upward trend, with only a modest overall improvement. Therefore, we adopt 128 outlier channels as our baseline setting.

\begin{table*}[h!]
\caption{The impact of different outlier channel number settings of the quantized model on the perplexity ($\downarrow$) and the zero-shot QA accuracy($\uparrow$). "FP16" denotes the performance of the original model represented in FLOAT16 format.}
\label{tab:outlierchannels}
\vskip 0.15in
\begin{scriptsize}
\begin{center}
\begin{tabular}{c|l|cccccccccc}
\toprule
\textbf{Model} & \textbf{Ch.} & \textbf{Wiki.}$\downarrow$ & \textbf{PTB}$\downarrow$ & \textbf{C4}$\downarrow$ & \textbf{PIQA}$\uparrow$ & \textbf{ARC-E}$\uparrow$ & \textbf{ARC-C}$\uparrow$ & \textbf{BoolQ}$\uparrow$ & \textbf{Hella.}$\uparrow$ & \textbf{Wino.}$\uparrow$ & \textbf{Avg.}$\uparrow$ \\
\midrule
\multirow{7}{*}{\textbf{\makecell{LLAMA1\\-7B}}} 
& FP16 & 5.68 & 27.34 & 7.08 & 77.37 & 52.48 & 41.38 & 73.06 & 72.99 & 67.01 & 64.05 \\
\cline{2-12}
& 0 & 471.19 & 1025.28 & 228.17 & 53.59 & 28.75 & 24.57 & 50.73 & 28.13 & 50.51 & 39.38 \\
& 128 & 8.58 & 76.09 & 12.27 & 68.88 & 45.03 & 30.89 & 69.63 & 55.41 & 59.35 & 54.87 \\
& 256 & 8.20 & 65.97 & 11.70 & 69.75 & 45.33 & 32.42 & 65.87 & 56.31 & 57.85 & 54.59 \\
& 512 & 7.80 & 57.44 & 10.90 & 71.27 & 47.94 & 34.22 & 65.57 & 58.46 & 59.19 & 56.11 \\
& 768 & 7.52 & 52.06 & 10.44 & 71.38 & 47.31 & 34.04 & 66.12 & 60.09 & 61.17 & 56.69 \\
& 1024 & 7.26 & 50.42 & 9.95 & 72.14 & 47.01 & 34.39 & 69.30 & 61.16 & 61.01 & 57.50 \\
\hline
\multirow{4}{*}{\textbf{\makecell{LLAMA2\\-7B}}} 
& FP16 & 5.47 & 22.51 & 6.97 & 76.93 & 53.58 & 40.53 & 71.07 & 72.96 & 67.17 & 63.71 \\
\cline{2-12}
& 128 & 8.89 & 69.46 & 12.74 & 68.72 & 46.13 & 30.55 & 66.12 & 55.76 & 58.01 & 54.22 \\
& 256 & 8.52 & 61.01 & 12.16 & 69.97 & 47.64 & 31.57 & 68.50 & 56.22 & 59.19 & 55.51 \\
& 512 & 8.00 & 56.77 & 11.43 & 69.80 & 46.97 & 31.66 & 67.83 & 57.55 & 60.77 & 55.76 \\
\hline
\multirow{4}{*}{\textbf{\makecell{Vicuna\\-v1.5-7B}}} 
& FP16 & 6.78 & 26.78 & 8.55 & 77.80 & 56.06 & 39.93 & 75.69 & 71.06 & 67.80 & 64.72 \\
\cline{2-12}
& 128 & 9.51 & 45.61 & 13.35 & 69.97 & 50.00 & 33.45 & 71.96 & 57.81 & 59.43 & 57.10 \\
& 256 & 9.28 & 44.01 & 12.94 & 70.35 & 51.05 & 34.04 & 72.97 & 58.42 & 61.40 & 58.04 \\
& 512 & 8.88 & 41.49 & 12.42 & 71.87 & 50.76 & 34.22 & 73.79 & 58.80 & 64.09 & 58.92 \\
\bottomrule
\end{tabular}
\end{center}
\end{scriptsize}
\vskip -0.1in
\end{table*}

\begin{figure*}[h!]
\vskip 0.2in
\begin{center}
\centerline{\includegraphics[width=\textwidth]{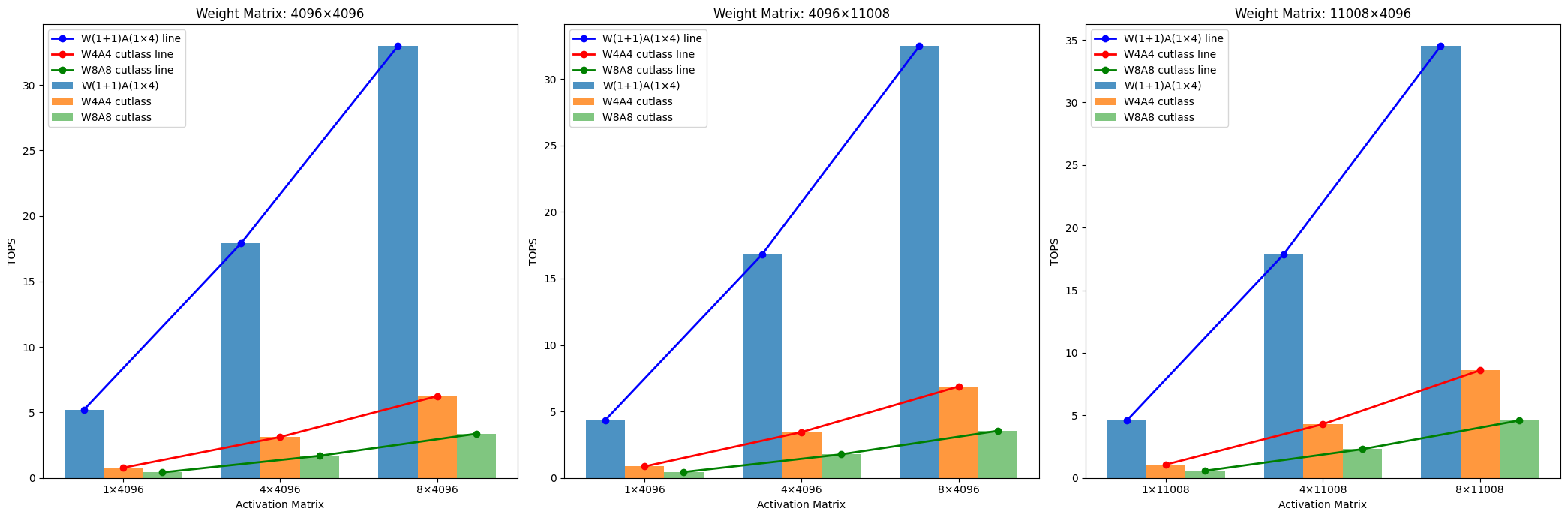}}
\caption{A comparison of the computational efficiency between the W(1+1)A(1×4) kernel and the INT4, INT8 kernels of CUTLASS for matrix multiplication with varying input lengths is conducted on the A6000.}
\label{pic:speed_full}
\end{center}
\vskip -0.2in
\end{figure*}

\smallskip\noindent\textbf{Comparison of different kernels.} In Figure \ref{pic:speed_full}, we comprehensively evaluate the performance of the W(1+1)A(1×4) kernel and the INT8, INT4 kernels from CUTLASS, based on the matrix multiplication sizes that may occur in the LLAMA model. Since our method incorporates a small amount of INT8 mixed-precision quantization, for the handling of outliers, we separately measure the computational efficiency of outliers and normal values. Subsequently, we derive the overall computational efficiency by considering the proportion of these two components.

\smallskip\noindent\textbf{Model Size.} We present in Table \ref{tab:gpumem} the theoretical compression effectiveness of our method on the LLAMA family. In our calculation of the model size, we have included both the quantization parameters and the additional storage incurred by fine-grained grouping, which results in our findings being slightly larger than those reported in BiLLM \cite{huang2024billm}. The binarization of weights significantly reduces the storage size of quantized LLMs and the GPU memory and bandwidth requirements during inference. Across LLAMA models of different sizes, our method achieves a compression ratio of over 5×. 

\section{AI Assistants in Research or Writing}
We use a local LLama3.3 model to polish the draft for checking grammar and improving expression. Research ideas, experiment design and discussion contents are all original by the authors.

\end{document}